\documentclass[11pt]{article}

\usepackage[colorlinks,urlcolor=blue,linkcolor=blue,citecolor=blue]{hyperref}
\usepackage{color,array}
\usepackage{graphicx}
\usepackage{amsmath}
\usepackage{tabularx}

\usepackage{bbold}
\usepackage{amssymb} 

\usepackage[margin=1in]{geometry}
\usepackage{fancyhdr}
\usepackage{comment}
\usepackage{parskip}
\usepackage{paralist}
\usepackage{siunitx}
\usepackage{hyperref}

\usepackage[skip=0.5\baselineskip]{caption}

\providecommand{\keywords}[1]
{
  \small        
  \textbf{\textit{Keywords---}} #1
}

\begin{document}

\title{Toward Reusability of AI Models Using Dynamic Updates of AI Documentation}

\author{Peter Bajcsy and Walid Keyrouz}

\maketitle

\begin{abstract}
This work addresses the challenge of disseminating reusable artificial
intelligence (AI) models accompanied by AI documentation (a.k.a., AI model cards). The work is
motivated by the large number of trained AI models that are not reusable
due to the lack of (a) AI documentation and (b) the temporal
lag between rapidly changing requirements on AI model reusability and
those specified in various AI model cards. 
Our objectives are to shorten the lag time in updating AI model card templates and align AI documentation more closely with current AI best practices.

Our approach introduces a methodology for delivering agile,
data-driven, and community-based AI model cards. We use the Hugging Face (HF) repository of AI
models, populated by a subset of the AI research and development
community, and the AI consortium-based Zero Draft (ZD) templates for the AI documentation
of AI datasets and AI models, as our test datasets. We also address questions about the value of AI documentation for AI reusability.

Our work quantifies the correlations between AI model downloads/likes
(i.e., AI model reuse metrics) from the HF repository and their
documentation alignment with the ZD documentation templates using tables
of contents and word statistics (i.e., AI documentation quality
metrics). Furthermore, our work develops the infrastructure to regularly compare AI documentation templates against community-standard practices
derived from millions of uploaded AI models in the Hugging Face
repository. The impact of our work lies in introducing a methodology for
delivering agile, data-driven, and community-based standards for
documenting AI models and improving AI model reuse.

\textbf{\textit{Impact Statement}}---Artificial Intelligence (AI) has become ubiquitous in our daily lives,
and its rapid advancements pose significant challenges for the reuse and
reproducibility of AI-based modeling. To harness the power of AI,
including building AI-based solutions on top of the AI investments
(labor, hardware, energy), and supporting AI integration into products,
AI model documentation (metadata about AI models) must be informative so
AI models can be reused by a broad community of consumers and suppliers.
However, there is a significant temporal lag between the best practices
in state-of-the-art AI-based solutions and the content of AI
documentation templates required for AI reuse. The impact of this work
lies in introducing a methodology and infrastructure to reduce lag in
updating AI documentation standards, aligning them more closely with
current best AI practices, and quantifying AI documentation quality.


\end{abstract}

\section{Introduction}
\label{sec:introduction}
This work is motivated by the large number of trained
artificial intelligence (AI) models that are not reusable due to
insufficient documentation and the lack of standards for the AI model documentation (a.k.a., AI model cards). The field of AI has been
advancing at an unprecedented pace~\cite{gil2025ai} that is outpacing the
traditional approaches to teaching new technology~\cite{brodheim2025shaping} and developing
standards needed for AI-based commerce. To harness the power of AI,
including building AI-based solutions on top of the AI investments
(labor, hardware, energy), and supporting AI integration into products,
AI model documentation (metadata about AI models) must be
informative for an AI model to be reusable by a broad community of
consumers and suppliers. This raises a question about the value of AI
model cards for AI model reuse, and what attributes of AI model cards are
important.

Among the numerous research and development (R\&D) challenges in
the fast-advancing AI field, the limited reuse of small- and
medium-sized AI models, as well as the reproducibility of AI-based
results, have become practical obstacles. 
The challenges can be attributed to lagging standards defining AI model
formats (ONNX file format effort~\cite{onnx2019}) and AI metadata documentation
required by integrators and consumers of AI-enabled products (e.g.,
developers and users of AI-enabled medical devices~\cite{fda2025ai}). This
implies that ``AI model card attributes'' must be agile and dynamically
adjusted as the field of AI advances.

\begin{figure}
  \centerline{\includegraphics[width=37pc]{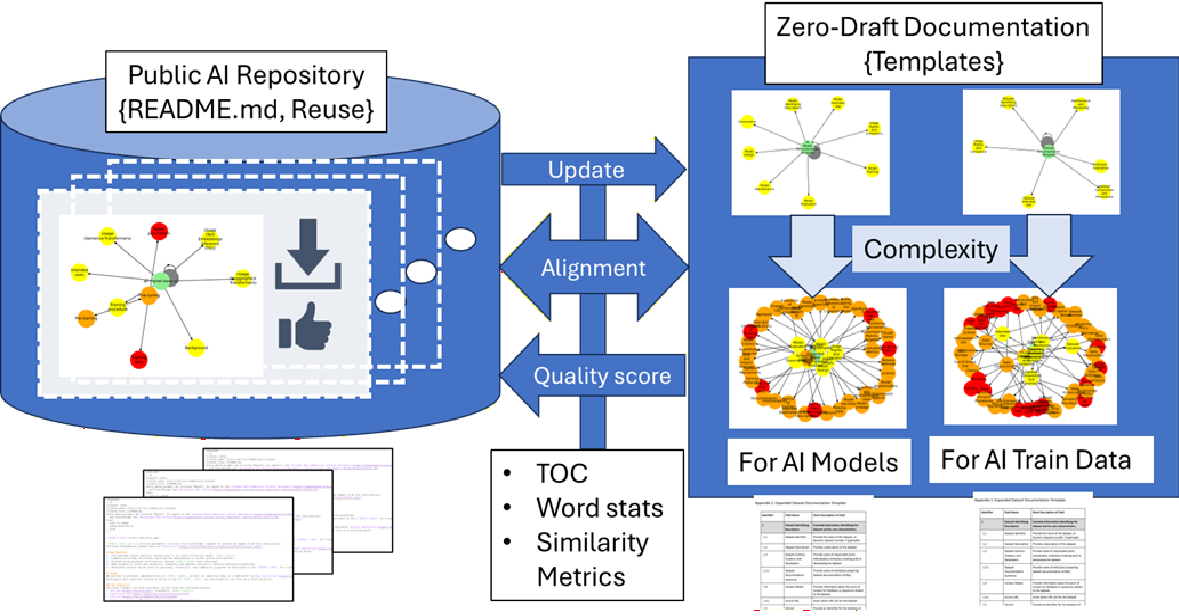}}
\caption{An overview of the interactions between public AI model repositories and Zero-Draft templates for AI models and AI training data.}
\label{fig:overview}

\centerline{\rule{0.6667\linewidth}{.2pt}}
\end{figure}

The problem of identifying attributes of AI documentation for rapidly
advancing AI technology lies in the temporal lag between the best
practices in state-of-the-art (SOTA) AI-based solutions and the content
of AI documentation templates required for AI model reuse. Our objectives
are to shorten the lag time, align the AI documentation templates
more closely with current AI best practices, and explore the
correlations between AI documentation quality and the reuse of AI
models. Our goal is to enable the interactions between public AI model
repositories and Zero-Draft AI documentation templates, as illustrated
in Figure~\ref{fig:overview}. We envision (a) the Zero-Draft templates to be updated
based on the best practices captured by the most downloaded/liked AI
models, (b) AI model contributors to public AI repositories receiving an
AI documentation quality metrics applied to Table of Contents (TOC) and word statistics, and 
(c) the implementation of a dynamic
mechanism to align community practices with reference AI documentation templates.

In our approach, we leverage the Hugging Face (HF) repository of AI
models~\cite{huggingface2026repo}, which is populated by a subset of the AI research and
development community, and the proposed Zero Draft (ZD) for a standard on
documentation of AI datasets and AI models~\cite{nist2025zero}, which is prepared by
the AI consortium, as our test datasets. Conceptually, our approach
is divided into two parts. First, we assess the value of AI
documentation for AI reusability by comparing HF subsets of README.md
files with high and low download counts against the ZD documentation
templates. The first part also serves as a prototype for scoring the quality
of AI documentation. Second, we compute a word histogram of all AI
documents in the HF repository and compare it with the word histogram of
the ZD documentation templates. The second part supports iterative, data-driven updates to ZD templates.

\section{Related Work}
\label{sec:related}

Related work to our approach can be classified into AI
documentation template recommendations by vendors and by AI communities.
First, AI documentation templates have been proposed by several research
teams~\cite{gebru2021datasheets, holland2018dataset, mitchell2019model} of information technology vendors
\cite{google2026model, facebook2026llama}, and AI model repository managers~\cite{huggingface2026cards, kaggle2026models}. The
template recommendations have been referred to as datasheets~\cite{gebru2021datasheets},
dataset nutrition label~\cite{holland2018dataset}, or model cards~\cite{mitchell2019model}. Second, the
community-based traditional approaches to developing documentation
standards leverage
\begin{inparaenum}[(a)]
\item community building via workshops and
\item preparations of Zero-Draft AI documentation templates based on discussions
during workshops and the help of small writing teams organized into
working groups, and
\item transitions of the Zero-Draft AI documentation templates to
the International Organization for Standardization (ISO). 
\end{inparaenum}
 The resulting
community-based recommendations require investments of time and human
resources, such as those by the current NIST AI Consortium~\cite{nist2025zero} or
the past ONNX community (the workshops have been discontinued, but the
online educational material is available~\cite{onnx2019}). Our work builds on AI
documentation recommendations from vendors, such as Hugging Face and
Google, and supports the iterative improvement of community-based AI
documentation. Our ultimate goal is to improve the reuse of
shared open and closed AI models~\cite{bajcsy2022characterization}. Our approach differs from the
past two approaches to developing standards, as it combines information
about AI best practices with community-driven AI documentation
templates.

The novelty of our work lies in enhancing the temporal agility of
AI documentation templates (a.k.a. Proposed Zero Draft for a Standard on
Documentation of AI Datasets and AI Models~\cite{nist2025zero}) and minimizing
the time and human resources required for frequent community workshops
to keep pace with rapid AI advancements. 
Our contributions lie in designing the methodology and software infrastructure to regularly
compare best practices derived from AI model cards in AI model repository and 
the SOTA ``standard'' for AI documentation templates. In
this case, we refer to ``best practices'' as community-common practices. Additionally, we can provide a simple
quality score for AI documentation relative to the reference documentation template at any time, 
helping AI developers improve the reusability of their AI models. Finally, we are answering a question about the value of aligning AI documentation with an AI model card reference for AI reusability via metrics capturing AI documentation structure and content. 

\section{Methodology}
\label{sec:methodology}

Our approach can be formulated as a set of research questions:

\begin{enumerate}
\item
  \emph{Table of contents alignment:} What is the similarity between the
  table of contents (TOC) recommended in the AI consortium-driven
  zero-draft (ZD) documentation templates~\cite{nist2025zero} and the TOCs used in the
  AI model cards uploaded to the Hugging Face (HF) repository~\cite{huggingface2026repo}?

\item
  \emph{Word histogram alignment:} What is the similarity between the
  word histograms derived from the AI consortium-driven zero-draft (ZD)
  documentation templates~\cite{nist2025zero} and from the AI model cards
  uploaded to the HF repository~\cite{huggingface2026repo}?

\item
  \emph{Correlation with AI model reusability:} Is there a correlation
  between the indicators of AI model reuse in the HF repository (number
  of downloads and likes) and the numerical similarity measurements
  established for TOCs and word histograms in the previous two research
  questions?

\item
  \emph{Data-driven updates of AI model card templates:} How to suggest
  dynamic updates to the AI consortium-driven ZD templates for training
  data and AI models based on the analyses of AI model cards in the HF
  repository?                                              
\end{enumerate}

\begin{figure}[t]
\centerline{\includegraphics[width=37pc]{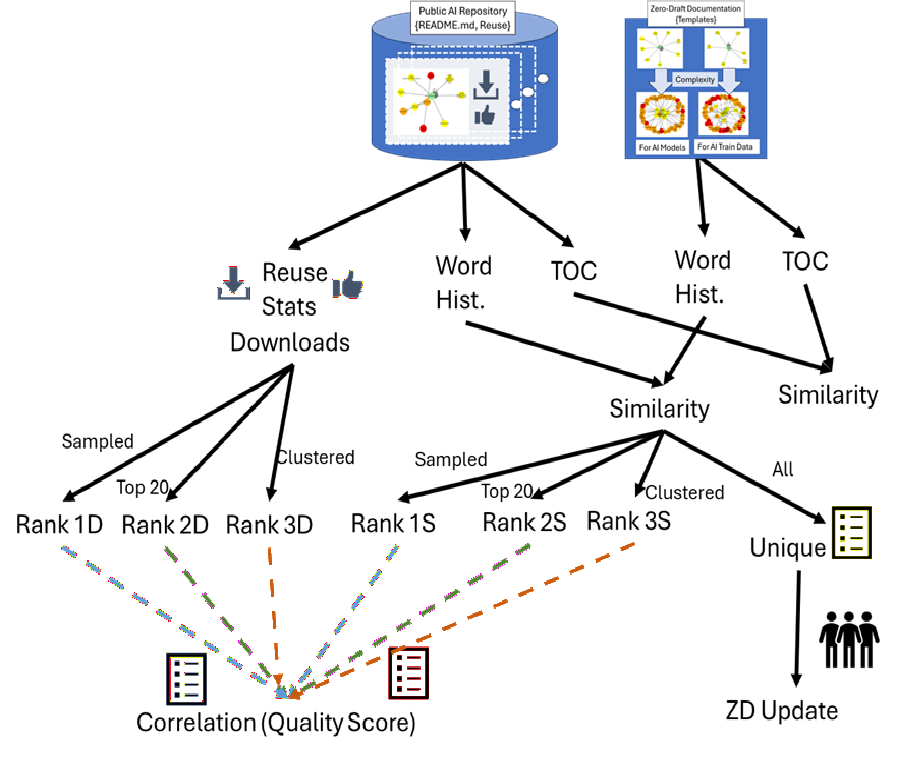}}
\caption{A workflow of steps to understand the alignment of AI documentation in public AI model repository (top left) and Zero-Draft templates (top right) by performing multiple similarity comparisons, rankings, correlations, and unique word analyses.}
\label{fig:workflow}

\centerline{\rule{0.6667\linewidth}{.2pt}}
\end{figure}

To answer the research questions, we designed an overarching workflow of
steps shown in Figure~\ref{fig:workflow}, where TOC stands for Table of Contents. This workflow allows us (a) to
report similarity metrics of AI model cards from the HF repository
\cite{huggingface2026repo} and the AI documentation templates from the Zero-Draft (ZD)
PDF file~\cite{nist2025zero}, (b) to compute correlations between similarity scores
of AI model cards with ZD data/model templates and the AI model
reusability indicators (number of downloads and likes), and (c) to make
numerically supported suggestions of ZD template updates. The correlations are computed between the ranks of AI models established based on the number of downloads (denoted as Rank XD in in Figure~\ref{fig:workflow}) and the ranks based on the word histogram similarity scores (denoted as Rank XS in Figure~\ref{fig:workflow}).  

\section{Metrics and Results}
\label{sec:methods-and-results}

We structured the work to focus on the raw AI model card dataset (Section \ref{sec:raw_dataset}), used subsets to assess the value of AI documentation
(Section \ref{sec:created_subsets}), tree representation of table of contents (Section \ref{sec:tree_rep}), and
similarity measurements of TOCs and word histograms computed over the
entire set of HF README.md files (Sections \ref{sec:sim_toc} and \ref{sec:sim_words}). The remaining two
sections address the problems of dynamic updates of ZD templates
(Section \ref{sec:update_sugg}) and quantification of AI model reusability as a function of
AI model card adherence to ZD templates (Section \ref{sec:correlations}). Each section
contains the metrics and the analysis of the AI model cards from the HF repository.

\subsection{Raw AI Model Card Dataset and Reference AI Model Card Templates}
\label{sec:raw_dataset}

In order to learn best practices in documenting AI models, we leverage
the Hugging Face (HF) repository of AI models~\cite{huggingface2026repo} because (a) it is the
largest open repository of AI models, (b) it is populated by the latest
models developed by the AI R\&D community, and (c) it provides a simple
application programming interface (API) for downloading AI
documentation. Table~\ref{table:dataset} summarizes the dataset creation based on the HF
repository API and the HF content as of 2025-12-22.

\begin{table}[t]
\centering
\caption{Statistics of the dataset downloaded from the Hugging Face (HF) AI model repository. README.md is the AI documentation in HF}
\label{table:dataset}


\begin{tabular}{|p{0.67\columnwidth}|p{0.28\columnwidth}|}
\hline
\textbf{Dataset Creation} & \textbf{Statistics} \\
\hline
Number of HF AI models (2025-12-22) & \num{2 305 649} AI models \\
\hline
Number of downloaded README.md files & \num{1 576 166} files \\
\hline
Number of AI models without README.md files (undocumented AI models) & \num{753 881} AI models \\
\hline
Estimated number of AI models with multiple README.md files in the HF directory structure & \num{24 398} AI models \\
\hline
Number of filtered README.md files if size \textgreater{} \SI{1}{\mega\byte} & 14 files (\SI{56.52}{\giga\byte}) \\
\hline
\textbf{Number of README.md files processed} & \textbf{\num{1 576 152}} files (\SI{4.92}{\giga\byte}) \\
  \hline
\end{tabular}

\vspace{2ex}
\centerline{\rule{0.6667\linewidth}{.2pt}}
\end{table}


The download of all \num{1 576 166} README.md files from the HF server took 17.87 days using a single
application request. This mapped to 0.49 seconds per request of the
README.md file. The value does not account for requests for which AI
models returned no results, nor for network fluctuations and variable
load on the HF server during the 18-day period. The files were
downloaded into 15 batches to enable parallel processing of the
downstream tasks.

We also downloaded the PDF file entitled ``Proposed Zero Draft for a Standard on
Documentation of AI Datasets and AI Models'' dated as of
September 2025~\cite{nist2025zero}. The PDF file was converted to text and trimmed
to include only Sections ``6.2 Dataset Documentation Template'', ``6.3 AI
Model Documentation Template'', and all appendices with template
extensions. These sections represent the structure and the keywords for
the community interested in AI documentation for training data and AI
models.

\subsection{Filtered HF and ZD Datasets and Subsets}
\label{sec:created_subsets}

\textbf{HF Set:} The downloaded HF files were filtered first. For example, the row in
Table~\ref{table:dataset} ``Number of filtered README.md files if size \textgreater{} 1
MB'' refers to 14 README.md files totaling 56.52 GB, which contain the
AI model coefficients instead of AI documentation, and therefore, they
were removed. Furthermore, we removed all Markdown language-specific
symbols (styling and structural elements, e.g., hash mark, three
hyphens, asterisks, or underscores), as they are relevant for extracting
AI documentation structure but are not included in semantic analysis. By
using word similarity metrics, we also removed the 198 stop words
defined by the Natural Language Toolkit (NLTK) for the English language
\cite{nltk2026}. While stop words are frequent, they carry minimal semantic
meaning and increase computational costs. Finally, we removed all
words/symbols in README.md files that contained more than two
occurrences of the letter x (\textbackslash x occurs in the hexadecimal
representation of binary characters). This removal addresses invalid
content in README.md files smaller than 1 MB, including the removal of
non-English characters and binary coefficients for smaller AI models
uploaded mistakenly as README.md. The rule of two occurrences of the
letter x is based on the observation that among about 5000 words defined in
the Oxford English Dictionary (OED) and 200 containing letter x \cite{OED2026}, none of the words contains more than two occurrences of x. 

For this HF Set, we sorted AI documentation files by the number of
downloads and the number of likes that are provided by the HF repository. 
The numbers of downloads and likes can be partially viewed as the HF metrics of AI model reusability and popularity. 

While we analyzed word histograms computed for the entire HF Set, we
assessed the value of AI documentation for smaller subsets. The subsets
also simplify visualizations of our results. We created three subsets of
all 1 576 152 AI model cards in the HF repository that had a README.md file (last row of Table~\ref{table:dataset}).
The model names and their download distributions are in shown in the Appendix~\ref{appendix:datasets}. 

\paragraph{HF Subset 1 (Uniform sampling of AI models.)} We sampled a sorted list of AI models by number of
downloads. We chose uniform sampling every 100,000th file, yielding a dataset containing 24 AI models, of which 8 lacked AI documentation. As a result, 16 AI models represent the number of downloads as follows: 148
550 580, 102, 30, 16, 12, 10, 8, 7, 6, 5, 2, and 0, and are shown in
Figure~\ref{fig:downloadsHFSet1}.

\paragraph{HF Subset 2 (Most reused AI models,)} We compile a dataset of the most popular AI models
by combining 21 models downloaded more than 10 million times and 20
models labeled with more than 4000 likes. This dataset contains 39 AI
models since two models met the download and likes thresholds
(sentence-transformers/all-MiniLM-L6-v2 and
meta-llama/Llama-3.1-8B-Instruct).

\paragraph{HF Subset 3 (Cluster-based sampling of AI models.)}
 We computed a histogram of all AI models based on the number of downloads. The histogram bins are 
increasing their width by a factor of 10 since the number of AI models with an increasing number of downloads is exponentially decreasing. The histogram is shown in Figure~\ref{fig:histdownloads}. We view the histogram as a method for download-based clustering of AI models. To explore the research hypothesis of a correlation between AI model documentation quality and AI model reuse, we sampled 20 README.md files per bin from five of the histogram bins with the number of downloads in the interval $[1, 100 000)$. This dataset contains 100 AI models and 8 of those 100 models do not have any documentation.

\textbf{ZD Set and Subsets:} The templates in the ZD document are
divided into training data card templates and AI model card templates.
We performed comparisons with the training data template (\textbf{ZD
Subset 1}), the AI model template (\textbf{ZD Subset 2}), and the
combined AI documentation (\textbf{ZD Set} = ZD Subset 1 $\bigcup$  ZD Subset 2).

\subsection{Tree Representation of Table of Contents}
\label{sec:tree_rep}

A structure of README.md can be extracted from the Markdown file format
by parsing it for the encoded multi-level headings, which represent the table of contents (TOC). 
We represent the TOCs as trees, with nodes corresponding to the headings
and edges connecting the root (the document) to its headings and
subheadings. 

\begin{figure}
\centerline{\includegraphics[width=37pc]{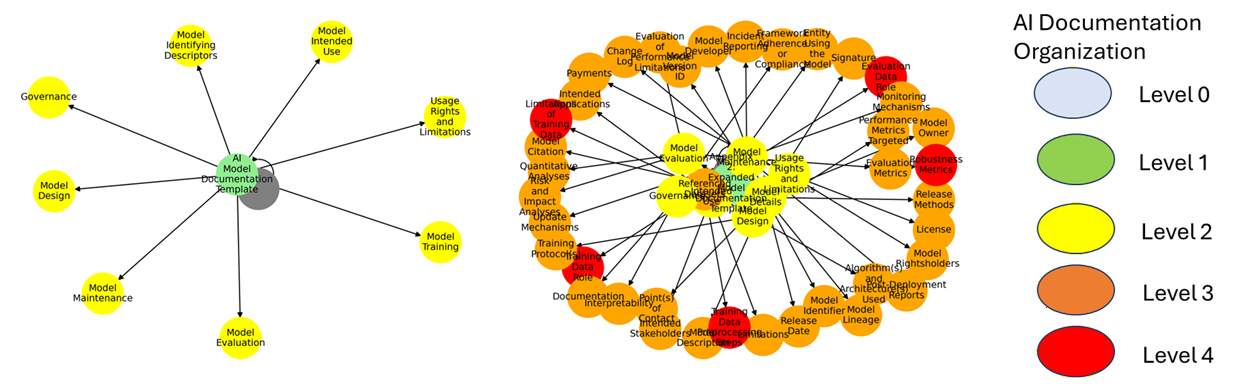}}
\caption{A tree representation of the ZD template for the highest-level headings of AI model cards (left) and for all levels of headings, including the Appendix of ZD (middle). The color mapping for the heading levels is shown on the right (gray is for the document root).}
\label{fig:zdtree}

\centerline{\rule{0.6667\linewidth}{.2pt}}
\end{figure}

Figure~\ref{fig:zdtree} shows the tree representation of the ZD template
for AI model cards. The figure conveys the complexity of the
specifications for the AI model card templates at the highest level
(left), as defined in the main document, and at the finest level
(right), as described in the Appendices of the ZD document~\cite{nist2025zero}. The complexity of the finest level
prevents us from showing legible headings in Figure~\ref{fig:zdtree}.

\begin{figure}[t]
\centerline{\includegraphics[width=37pc]{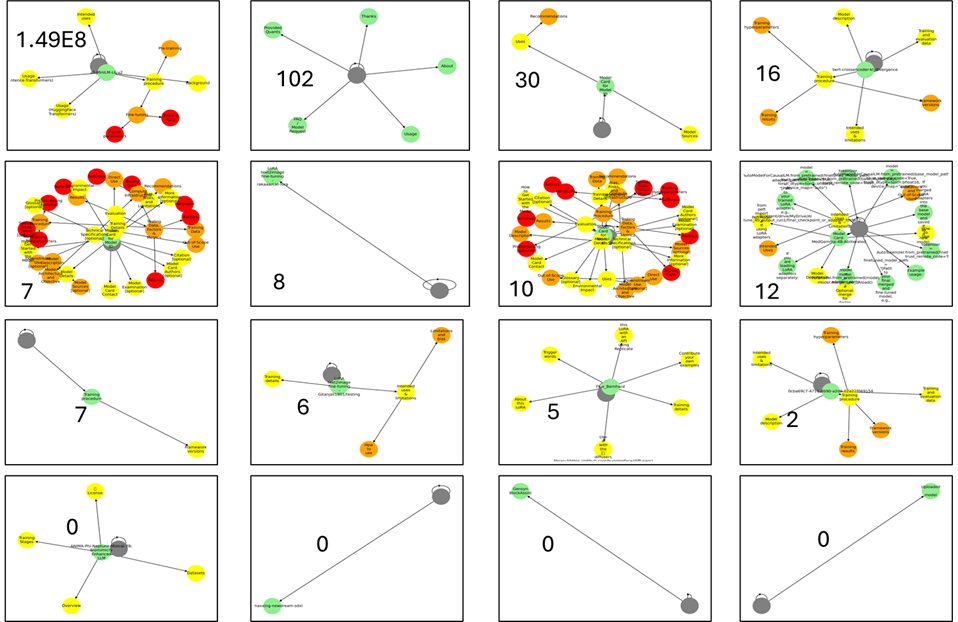}}
\caption{TOC trees for the 16 AI model cards in HF Subset 1. The overlaid numbers correspond to the number of downloads. }
\label{fig:tochfsubset1}

\centerline{\rule{0.6667\linewidth}{.2pt}}
\end{figure}

Figure~\ref{fig:tochfsubset1} shows 16 tree representations of the TOCs extracted from the README.md
files contained in the HF Subset 1. 
The color mapping shows green, yellow, orange and red as heading levels 1, 2, 3, and 4 in this order.
The trees are overlaid with the
number of downloads per model, and sorted by the number of downloads. 
Based on the visual inspection of Figure~\ref{fig:tochfsubset1}, the tree complexity and the number of AI model downloads are uncorrelated. Additional visualizations of TOCs extracted from the README.md files in HF Subsets 2 and 3 can be found in the Appendix~\ref{appendix:toc}.

\subsection{Similarity Metrics Applied to Table of Contents}
\label{sec:sim_toc}

To compare headings in TOCs of AI models cards from HF with the ZD
templates, we explored metrics such as Longest Common Subsequence (LSS)
and multiple versions of Levenshtein Distance (LD)~\cite{yao2023quality, behara2020novel}
applied to two strings. In our case, the TOC tree structure is converted
into a set of strings, each containing the words from the parent node to
each child tree node (e.g., Parent = ``Evaluation'', Child 1 = ``Metric'', Child 2 = ``Speed'' is converted to three strings for the three nodes: ``Evaluation'', ``Evaluation Metric'', and ``Evaluation Speed''). 
This results in the number of ``heading'' strings
equal to the number of tree nodes.

We computed the normalized LSS (NLSS) distance according to Equation
\eqref{eq:nlss}, where s(HF) and s(ZD) are two strings representing the headings
from each HF model card and ZD templates, and
$\left( s(HF) \cap s(ZD) \right)$ refers to a set of common words. We
first establish heading string matches if the NLSS value is larger than
$Thresh_{NLSS\ MATCH}= 25$ to ensure that at least one quarter of the
heading length in HF documentation matches exactly one of the headings
in the ZD template. Next, we count the number of HF AI model cards that
had at least one heading match as a measure of compliance with the ZD
template (the lowest bar for compliance).

\begin{equation}
\label{eq:nlss}
\begin{aligned}
v1 &= \text{length}\left( s(HF) \cap s(ZD) \right) \\
v2 &= \text{length}\left(s(HF)\right) \\
NLSS &= 100 \cdot \frac{v1}{v2}
\end{aligned}
\end{equation}

The Levenshtein distance (LD) is calculated between the sorted tokens of
the two input strings and calculates the least expensive set of
insertions, deletions, or substitutions that are needed to transform
s(HF) to s(ZD). Among all versions of the Levenshtein distance, such as
ratio, partial ratio, token sort ratio, and token set ratio available in
the Python FuzzyWuzzy library, we used the normalized ratio
($NLD_{ratio}$) and token sort ratio ($NLD_{sorted}$) comparisons
defined in Equation \eqref{eq:gld_nld} (for more information, see Behara et al.
\cite{behara2020novel}). In Equation \eqref{eq:gld_nld}, GLD is the generalized LD defined as the
minimum cost associated with deletion, insertion, and substitution
operations applied to characters in one string to convert it to another
string, and s is the sequence of operations to match s(HF) to s(ZD). The
normalization factor is the sum of the lengths of the two compared
strings. The difference between $NLD_{ratio}$ and $NLD_{sorted}$
lies in sorting the words in strings for $NLD_{sorted}$ before the
deletion, insertion, and substitution operations are applied.

\begin{equation}
\label{eq:gld_nld}
\begin{aligned}
GLD &=  \min_{S}{\sum_{k = 0}^{s}{\text{cost}(\text{operation}_{k})}} \\
NLD_{ratio} &=  \frac{GLD}{\left| s(HF) \right| + \left| s(ZD) \right|}
\end{aligned}
\end{equation}

Similar to the NLSS similarity metric, we first declare a match of
headings if the $NLD_{sorted}$ value is larger than
$Thresh_{NLD\ MATCH}= 50$ to ensure that after the sequence operations
are applied, at least one half of the heading in the HF documentation
matches one of the headings in the ZD template. Next, we count the
number of HF AI model cards that had at least one heading match as a
measure of compliance with the ZD template (the lowest bar for
compliance).

Finally, we compute the similarity of TOCs using the NLSS and NLD metrics for the HF Subsets ${j} \in \{ 1,2,3\}$ according to Equation~\ref{eq:percent}, where $n_{j}$ is the number of README.md files in each HF Subset.
\begin{equation}
\label{eq:percent}
\begin{aligned}
Sim_{j}^{NLSS}=\frac{1}{n_{j}}  \sum_{i=1}^{n_{j}} \mathbb{1}_{\{NLSS_{i} \geq 25 \}} \\
Sim_{j}^{NLD}=\frac{1}{n_{j}}  \sum_{i=1}^{n_{j}} \mathbb{1}_{\{NLD_{i} \geq 50 \}}
\end{aligned}
\end{equation}


\paragraph*{Results}
We provided examples of NLSS and NLD similarity metrics that are applied to a single README.md file 
in Appendix~\ref{appendix:similarity}. The examples of NLSS and NLD similarity metrics are useful 
for our understanding of the
correlation between the verbosity of headings and the relevant
information as defined in the ZD templates.

Given the two similarity metrics, Tables~\ref{table:nlss_similarity2} and ~\ref{table:nld_similarity2} present the numerical
comparisons of NLSS- and NLD-based comparisons of TOCs in ZD Subsets 1
and 2 (training data and AI model templates) and TOCs in HF Subsets 1, 2, and 3.
While the thresholds for NLSS and $NLD_{sorted}$ were selected
empirically, both tables provide consistent evidence that the AI model
cards from HF Subset 2 (HF top downloads \& liked) are more aligned with
the ZD Subsets 1 and 2 (data and model templates) than the AI model
cards from HF Subsets 1 and 3 (uniformly sampled and clustered). 
The alignment differences between HF Subset 2 (the most popular) and HF Subset 3 (excluded the most popular and undocumented) are  
21.63\% (ZD data) and 19.07\% (ZD model) for the NLSS-based comparison, and 
15.25\% (ZD data) and 14.24\% (ZD model) for the $NLD_{sorted}$-based comparison.
These numbers indicate the potential for improving AI documentation with respect to the documentation of the most reused AI models. The potential for improving the documentation of training data is slightly larger than AI models, which reflects that the developers document more the attributes of AI models than the data used for training those disseminated AI models. 


\begin{table}[t]
\centering
\caption{NLSS-based comparisons of TOCs from HF Subset 1, 2, and 3 against the two types of templates in ZD templates (data and AI model). The similarity values are computed according to Equations \eqref{eq:nlss} (per file) and \eqref{eq:percent} (per HF subset).}
\label{table:nlss_similarity2}
\begin{tabularx}{\columnwidth}{|>{\raggedright\arraybackslash}X|>{\centering\arraybackslash}X|>{\centering\arraybackslash}X|}
\hline
\textbf{Percent of HF files for which min headings: NLSS$\geq 25$} & \textbf{ZD Data Template (ZD Subset 1)} & \textbf{ZD Model Template (ZD Subset 2)} \\ \hline
HF Subset 1 (16)& 31.25 & 43.75  \\ \hline
HF Subset 2 (39)& 56.41 & 53.85  \\ \hline
HF Subset 3 (92)& 34.78 &       34.78  \\
\hline
\end{tabularx}

\vspace{2ex}
\centerline{\rule{0.6667\linewidth}{.2pt}}
\end{table}


\begin{table}[htbp]
\centering
\caption{$NLD_{sorted}$-based comparisons of TOCs from HF Subset 1, 2, and 3 against the two types of templates in ZD templates (data and AI model). The similarity values are computed according to Equations \eqref{eq:gld_nld} (per file) and \eqref{eq:percent} (per HF subset).}
\label{table:nld_similarity2}
\begin{tabularx}{\columnwidth}{|>{\raggedright\arraybackslash}X|>{\centering\arraybackslash}X|>{\centering\arraybackslash}X|}
\hline
\textbf{Percent of HF files for which min headings: $NLD_{sorted} \geq 50$} & \textbf{ZD Data Template (ZD Subset 1)} & \textbf{ZD Model Template (ZD Subset 2)} \\ \hline
HF Subset 1 (16) & 31.25        & 31.25 \\ \hline
HF Subset 2 (39) & 35.9 & 43.59 \\ \hline
HF Subset 3 (92) & 20.65        & 29.35 \\
\hline
\end{tabularx}

\vspace{2ex}
\centerline{\rule{0.6667\linewidth}{.2pt}}
\end{table}

\subsection{Similarity Metrics Applied to Word Histograms}
\label{sec:sim_words}

Word histograms are a fine-grained representation of AI model cards in
comparison to TOCs. Word histograms change over time as new aspects of
AI model reuse are discovered and captured in AI model cards. 
The histogram was computed
for all AI model cards in the HF repository, across downloaded 15 batches of
README.md files. For the word histogram, the
15 partial histograms were merged into a single histogram. The total CPU
time for the word histogram computation was 31.94 hours on a Dell laptop
(Windows 11, 11th Gen Intel(R) Core(TM) i7-11850H @ 2.50 GHz). The same
histogram calculation was applied to the templates defined in 
the Zero Draft dated as of September 2025~\cite{nist2025zero}. The total
number of unique words in the histogram derived from the HF repository
is 841 220, and from the Zero Draft is 305. Both histograms are plotted in Appendix~\ref{appendix:hist}.

Let us assume that the word histogram of ZD templates (ZD Set) contains
all the necessary description words needed to reuse an AI model. Then we
can compare two computed word histograms using the following metrics:

\begin{enumerate}
\item
  Number of common words (count of words in Venn diagram intersection)
\item
  Histogram intersection (sum of frequencies for common words)
\item
  Cosine similarity: 
\begin{equation}
\label{eq:cosine}
\cos (H1, H2) = \frac{ \sum [H1(w) \cdot H2(w)] } { \sqrt{ \sum [H1(w)]^2 } \cdot \sqrt{ \sum [H2(w)]^2 } }
\end{equation}
  where H(w) is the
  frequency of a word w in the histogram of ZD templates (H1(w)) or HF
  AI model cards (H2(w)) and the sums are taken over all common words w
  in the two vocabularies
\item
  KL Divergence: 
\begin{equation}
\label{eq:kld}
KLD(H1 \parallel H2) = \sum \left( \frac{H1(w)}{\sum H1(w')} \cdot \log \frac{ \frac{H1(w)}{\sum H1(w')} }{ \frac{H2(w)}{\sum H2(w')} } \right)
\end{equation}
  where each H(w) is normalized to
  a probability by the sum of word frequencies in the H vocabulary. Note
  that KL Divergence is asymmetric 
\begin{equation}
\label{eq:kld_asym}
KLD(H1 \parallel H2) \neq KLD(H2 \parallel H1)
\end{equation}
  and therefore, we consider either the histogram of ZD
  templates (H1(w)) or the histogram of HF AI model cards (H2(w)) as a
  reference.
\end{enumerate}


\paragraph*{Results}
Table~\ref{table:metrics} includes numerical values of the above metrics as well as the
counts of words that are unique to ZD templates or to HF AI model cards.
The cosine similarity value suggests that the vector of words in the ZD
template represents about 45° alignment with the vector of words in the
HF histogram (the same vector direction). The KL divergence of the HF
word histogram from the ZD word histogram (15.164) is about three times
larger than the KL divergence of the ZD word histogram from the HF word
histogram (5.751), which indicates a distance gap between the two
distributions (i.e., there is three times more information lost when a
ZD histogram is used to approximate a HF histogram than vice versa).

Table~\ref{table:uniquewords} shows the list of 28 unique words in ZD templates
(count\_ZD\_only\_words in Table~\ref{table:metrics} from the ZD Set). These words are
never used in HF AI model cards and require further investigation into
the reasons. For instance, ``reliability'' is not specified in AI model
cards because it is an open research problem, while ``addendums'' are
never included in AI model cards prepared by developers.


\begin{table}[t]
\centering
\caption{A summary of metrics used to compare the word histograms extracted from ZD templates (ZD Set) and from all HF AI model cards, which are summarized in Table 1.}
\label{table:metrics}
\begin{tabularx}{\columnwidth}{|>{\raggedright\arraybackslash}X|>{\centering\arraybackslash}X|}
\hline
\textbf{Metric name} & \textbf{Value} \\ \hline
count\_common\_words & 277 \\ \hline
count\_ZD\_only\_words & 28 \\ \hline
count\_HF\_only\_words & 840 943 \\ \hline
histogram\_intersection & 1462 \\ \hline
cosine\_similarity & 0.447 \\ \hline
Reference=ZD: kl\_divergence & 15.164 \\ \hline
Reference=HF: kl\_divergence & 5.751 \\
\hline
\end{tabularx}

\vspace{2ex}
\centerline{\rule{0.6667\linewidth}{.2pt}}
\end{table}


\begin{table}[htbp]
\centering
\caption{A list of unique words in ZD templates (ZD Set) with respect to all HF AI model cards.}
\label{table:uniquewords}
\begin{tabularx}{\columnwidth}{|>{\raggedright\arraybackslash}X|>{\centering\arraybackslash}X|>{\centering\arraybackslash}X|>{\centering\arraybackslash}X|}
\hline
\textbf{word} & \textbf{frequency} & \textbf{word} & \textbf{frequency} \\ \hline
accessible & 2 & capabilities & 3 \\ \hline
behavior & 2 & available & 3 \\ \hline
being & 2 & bundle & 3 \\ \hline
box & 2 & which & 3 \\ \hline
obtained & 2 & attributions & 3 \\ \hline
reliability & 2 & describing & 4 \\ \hline
substituted & 2 & feedback & 4 \\ \hline
addendums & 2 & applicable & 4 \\ \hline
robustness & 2 & rightsholders & 4 \\ \hline
base & 2 & benchmarks & 5 \\ \hline
vocabulary & 2 & readable & 6 \\ \hline
publicly & 3 & distribution & 7 \\ \hline
number & 3 & subset & 14 \\ \hline
responsible & 3 & describe & 40 \\
\hline
\end{tabularx}

\vspace{2ex}
\centerline{\rule{0.6667\linewidth}{.2pt}}
\end{table}

\subsection{Update Suggestions}
\label{sec:update_sugg}

\begin{figure}[t]
\centerline{\includegraphics[width=37pc]{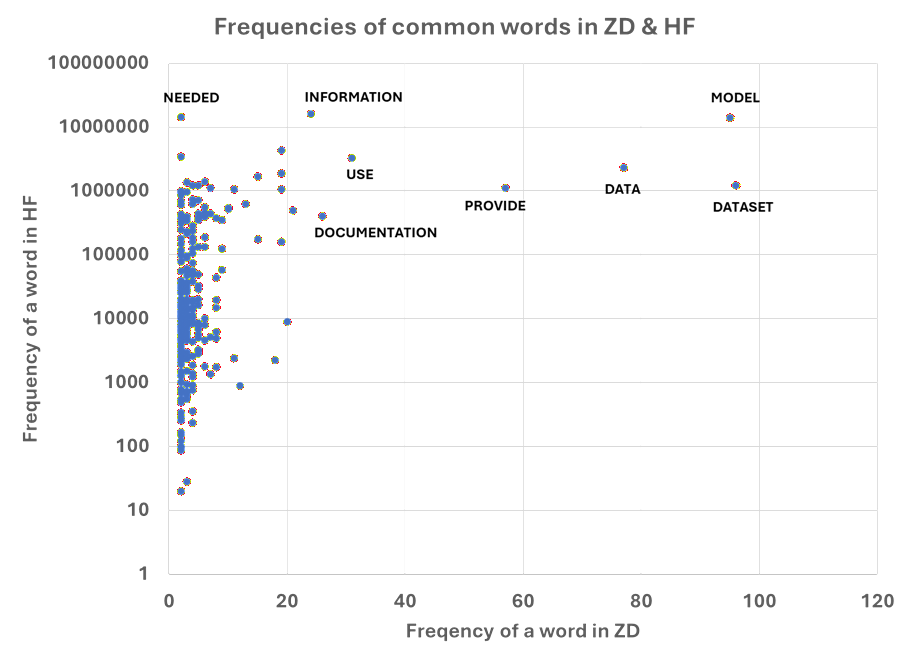}}
\caption{Scatter plot of common words in the Venn diagram of HF and ZD word histograms.
}
\label{fig:scatter}

\centerline{\rule{0.6667\linewidth}{.2pt}}
\end{figure}

The similarity of word histograms in Figure~\ref{fig:workflow} separates words from the two
input histograms into three categories: common words, only HF words, and
only ZD template words. 
Figure~\ref{fig:scatter} shows a scatter plot of common words in the Venn diagram of HF and ZD word histograms. We placed words only at the points with high frequencies of occurrence in ZD or HF word histograms. 

We listed the top 20 most frequently counted
words from each histogram in Table~\ref{table:commonwords} (right-most columns). The words in
Table~\ref{table:commonwords} can be filtered to eliminate redundancies, for instance, verbs
and nouns (used -- use, describe-description), relationships
(dataset-subset), and possible synonyms (results, summary). 
For the
update suggestions, one should focus on the words that occur only in one
of the columns in Table~\ref{table:commonwords} because of their high frequency and
uniqueness.

\begin{table}[t]
\centering
\caption{Comparison of word sets. HF = Hugging Face, ZD = AI Zero Draft}
\label{table:commonwords}
\begin{tabularx}{\columnwidth}{|>{\raggedright\arraybackslash}X|>{\centering\arraybackslash}X|>{\centering\arraybackslash}X|>{\centering\arraybackslash}X|}
\hline
\textbf{Word (top 20 in common words sorted by $f(ZD) * f(HF)$)} & \textbf{Word (top 20 in ZD)} & \textbf{Word (top 20 in HF)} \\ \hline
model & dataset & information \\ \hline
information & model & needed \\ \hline
data & data & model \\ \hline
dataset & provide & optional \\ \hline
use & describe & training \\ \hline
training & use & section \\ \hline
provide & documentation & use \\ \hline
evaluation & information & card \\ \hline
needed & including & data \\ \hline
limitations & identify & evaluation \\ \hline
description & description & precision \\ \hline
metrics & example & mixed \\ \hline
documentation & evaluation & limitations \\ \hline
including & training & nan \\ \hline
technical & organizations & used \\ \hline
intended & limitations & log \\ \hline
link & performance & technical \\ \hline
section & subset & relevant \\ \hline
results & intended & summary \\ \hline
license & analyses & dataset \\ 
\hline
\end{tabularx}

\vspace{2ex}
\centerline{\rule{0.6667\linewidth}{.2pt}}
\end{table}

To understand the gap between ZD and HF word histograms, Table~\ref{table:commonwords} (left
column) also shows the top 20 words from the common word set of 277
words, sorted by multiplying the frequencies of common words across the
ZD and HF word histograms. The $f(ZD)*f(HF)$ metric can reveal insights
about what keywords are under-represented in the ZD templates (low value
of $f(ZD)$ but highly used in HF AI model cards (high value of $f(HF)$). In
the ZD templates, the emphasis is on ``identify'', ``organizations'',
``performance'', ``analysis'', and ``example''. In the HF data source,
the emphasis is on ``card'', ``relevant'', ``mixed'', ``precision'',
``log'', ``NaN'', and ``optional''. One of the many interpretations of
the differences lies in the distinction between specific and abstract
descriptions of key AI model attributes (e.g., the abstract description
``performance analysis'' in ZD versus the specific description ``mixed
precision'' in the HF data source). Another interpretation of the
differences is the focus on attribution in ZD versus easy-to-run aspects
in HF (e.g., ``organizations'' versus ``log'', ``NaN''). Unique words
selected based on f(ZD)*f(HF), such as ``metrics'', ``link'', and
``license'', are also important in AI model cards according to both HF
and ZD data sources. The ZD templates might benefit from more details
about metric definitions, license types, and links to additional
information resources.

\subsection{Correlations of Similarity Metrics and Reusability Indicators}
\label{sec:correlations}

With our focus on AI model reuse through improved documentation, we
consider the number of downloads of AI models as
quantitative indicators of reusability. To quantify the importance of AI
model card quality for AI model reuse, we correlate their ranks
computed based on the similarity to the ZD templates (higher similarity implies smaller rank value) 
and the ranks of AI models based on reusability indicators (higher number of downloads implies smaller rank value).
Table~\ref{table:correl} summarizes the correlation values for the three HF Subsets. ``Rank vs Rank'' refers to correlations of README.md files in a subset that are ranked by common words versus by number of downloads. 
``Freq vs Freq'' implies that the correlations are computed over pairs of frequencies of words and downloads in each subset of README,md files.

\begin{table}[t]
\centering
\caption{A summary of correlation results.}
\label{table:correl}
\begin{tabularx}{\columnwidth}{|>{\raggedright\arraybackslash}X|>{\centering\arraybackslash}X|>{\centering\arraybackslash}X|}
\hline
\textbf{Correlation} & \textbf{Rank vs Rank} & \textbf{Freq vs Freq} \\ \hline
HF Subset 1 & 0.31 & 0.23 \\ \hline
HF Subset 2 & -0.04 & -0.06 \\ \hline
HF Subset 3 & 0.33 & 0.15 \\ \hline
\end{tabularx}

\vspace{2ex}
\centerline{\rule{0.6667\linewidth}{.2pt}}
\end{table}

\begin{figure}[t]
\centerline{\includegraphics[width=37pc]{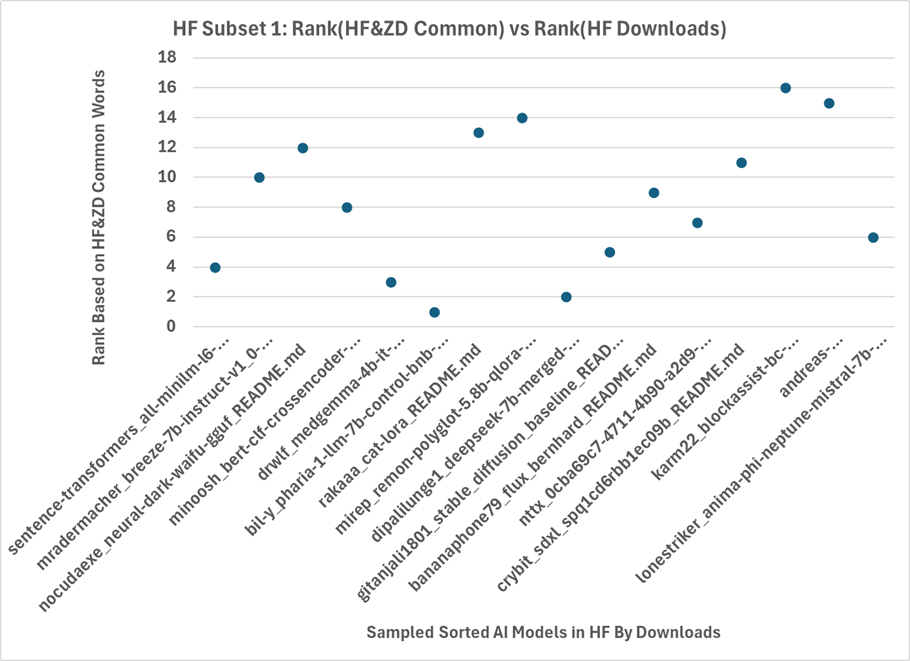}}
\caption{A scatter plot of ranks of HF AI model cards from HF Subset 1 according to the number of downloads (horizontal axis) and the common word similarity (vertical axis). The ranks are sorted from the largest to the smallest number of downloads (left-to-right) or similarity values (bottom-up).}
\label{fig:rank1}

\centerline{\rule{0.6667\linewidth}{.2pt}}
\end{figure}

\paragraph*{HF Subset 1} Figure~\ref{fig:rank1} illustrates the ranking of AI model cards in Subset 1, defined
in Section \ref{sec:created_subsets} (16 AI models in HF with a variable number of downloads).
The ranks are computed based on the count of overlapping words with the
word histogram of the ZD AI templates (i.e., common words). The smaller
rank value reflects a higher number of overlapping words and better adherence
to the ZD templates. Figure~\ref{fig:rank1} shows AI models sorted by 
the decreasing number of downloads from left to right along the
x-axis. We observe an increasing trend in rank with fewer downloads
(less adherence to the ZD templates implies a higher similarity-based rank value, 
the slope of $\log(\textrm{downloads})=f(\textrm{common words})$ is $3.39$). 

One could replace the ranking by the common words in ZD and HF histograms with the all words in the HF histogram and compute the common words for each README.md file.
Appendix~\ref{appendix:correlation} includes such a ranking of AI model cards for Subset 1.

\paragraph*{HF Subset 2} We quantified the correlations of the number of
downloads in Subset 2 (21 AI models in HF with the top number of downloads) and their frequencies of
overlapping words with the ZD word histogram (i.e., common words). The
correlation values (``Rank vs Rank'' is $-0.04$ and ``Freq vs Freq'' is$-0.06$) illustrates that the most reusable AI
models exhibit very similar adherence to the ZD templates.

\begin{figure}[t]
\centerline{\includegraphics[width=37pc]{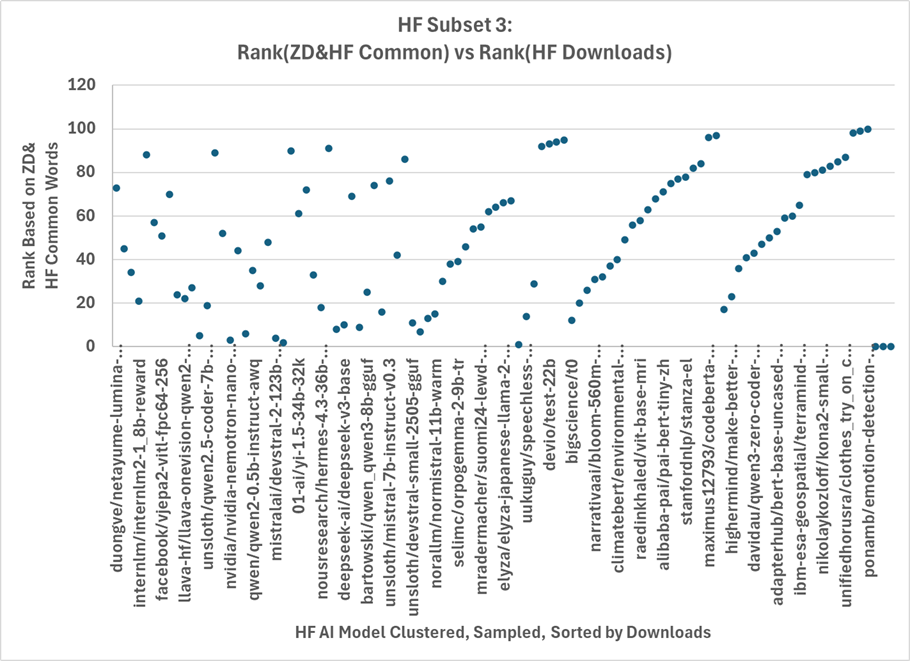}}
\caption{A scatter plot of ranks of HF AI model cards from HF Subset 3 according to the number of downloads (horizontal axis) and the common word similarity (vertical axis). The ranks are sorted from the largest to the smallest number of downloads (left-to-right) or similarity values (bottom-up).}
\label{fig:rank3}

\centerline{\rule{0.6667\linewidth}{.2pt}}
\end{figure}

\paragraph*{HF Subset 3} Figure~\ref{fig:rank3} displays a similar ranking of AI model cards in Subset 3 to
Figure~\ref{fig:rank1}. The HF Subset 3 consists of README.md files that represent samples of ``typical'' reusable AI models 
(number of downloads larger than $1$ and less than $10^5$).
The ``Rank vs Rank'' correlation value $0.33$ for Subset 3 is similar to the correlation value $0.31$ for Subset 1, but the ``Freq vs Freq'' correlation values are dissimilar ($0.23$ vs $0.15$). We hypothesize that the frequencies of key words in AI documentation may relate to reusability of AI models.

\section{Discussion}
\label{sec:discussion}

\paragraph*{Value of AI documentation} The numerical correlations 
between AI documentation quality and AI model
reuse in Table~\ref{table:correl} 
indicate some value in preparing high quality AI model cards.
The correlation magnitude can be interpreted
through the multi-faceted ranking of AI models during community
adoption, which encompasses factors such as AI model performance,
specific hardware and software requirements for running a model, and the
quality of documentation (among others). While there is some correlation
to AI documentation quality across the widely varying AI models in the
sorted, clustered, and sampled sets (Subsets 1 and 3), no such dependence was observed among
the most popular AI models in the HF repository (Subset 2). This can be
interpreted as adequate AI documentation investments by all authors of
these popular AI models, compared to other investments in building their
models, which enable reuse at scale with 10 million users or more.

\paragraph*{Reusability of AI models} Since the quality of the AI model card is just one of many contributing
factors to AI model reusability, the correlations and multiple
comparison metrics must be applied to the actual re-execution of the
downloaded AI models from HF. In this work, we excluded the sections of
AI model cards dedicated to setting up a virtual environment, installing
all necessary libraries, downloading test data, running the AI model on
the test data, and comparing the inferences with the expected outcomes.
These steps must be automated based on the AI model cards, which will
provide a real test of the quality of AI model documentation. One can
develop an agentic large language model for such a solution with a set
of tools and secured hardware.

\paragraph*{Limitations} The infrastructure for integrating best practices and community-defined
AI documentation fields are fine-tuned to English characters and
address some of the current issues of erroneous uploads of AI models
in place of AI documentation. However, robustness against all erroneous
uploads is not guaranteed at scale, with millions of users uploading
millions of AI models. The multilingual contributions to the HF
repository pose a challenge in constructing a multilingual ZD template.
Finally, the Venn diagram sets depend on the word match definitions, as
words can have many semantically similar forms (such as conjugations,
singular versus plural, etc.). In our work, we used exact word matching
when computing histograms, which could have biased the metrics. We would
also note that reusing AI models is very difficult if the models lack a
pipeline tag indicating their task. The current number of AI models that
have None as the pipeline tag is 1 519 249.

\paragraph{AI model card updates} 
Human-in-the-loop is necessary for the ZD Update step shown in Figure~\ref{fig:workflow}. 
One would need to run multiple analyses of word
histograms and word co-occurrence histograms over time to establish
patterns and rules for fully automated updates of ZD templates. The
protocol for revising the AI model documentation templates must follow
the governance and technical reviews. The automation of this step is
currently unknown and could be investigated after collecting data about
the template updates across several iterations over time.

\section{Conclusion}
\label{sec:conclusion}

We introduced a methodology for updating reference AI
documentation templates (also known as AI model cards) to improve the
reusability of AI models. We developed the infrastructure to compare AI
documentation templates (Zero-Draft AI templates for training data and
AI models) against community best practices for documenting AI models
uploaded to the Hugging Face repository. Finally, we assessed the value
of AI documentation for AI model reuse.

In this work, we have illustrated that TOC complexity is uncorrelated
with the number of downloads in Section \ref{sec:tree_rep}. 
The verbosity of TOC headings
might not imply better numerical alignment with the ZD templates shown
in Section \ref{sec:sim_toc}. 
We also concluded in Section \ref{sec:sim_toc} that if NLSS- and
NLD-based comparisons of TOCs in ZD data and model templates with the HF
Subset 2 (the most popular AI models) are viewed as a baseline for good AI documentation, 
then a typical AI model card in HF Subset 3 can be improved by around 20\% (NLSS metric) or 15\% (NLD metric) for the choices of NLSS and NLD thresholds.
While the total number of unique words derived from the HF repository is 841 220 and from the
Zero Draft is 305, their cosine similarity suggests that these vectors of unique words have about 45° alignment, and their KL divergence indicates three times more information lost when a ZD histogram is used to approximate a HF histogram than vice versa (Section~\ref{sec:sim_words}). 
To understand the gap between ZD and HF word histograms, we
introduced the $f(ZD)*f(HF)$ metric in Section \ref{sec:update_sugg}, which can provide
insights into which keywords are under-represented in the ZD templates
(low value of $f(ZD)$, but highly used in HF AI model cards (high value of
$f(HF)$). Finally, to quantify the importance of AI model card quality for
AI model reuse, we estimated the adherence of HF README.md files to the
ZD templates in Section \ref{sec:correlations} via word histogram overlap. We concluded
that there is a weak correlation of ``Rank vs Rank'' and ``Freq vs Freq'' for HF Subsets 1 and 3, 
when the common words between the ZD templates and
all HF AI model cards are used as a reference for Rank and Freq.
There is no correlation for HF Subset 2, which illustrates that the most reusable AI
models in Subset 2 exhibit very similar adherence to the ZD templates,
whereas AI models with widely varying reusability demonstrate varying
quality AI model documentation.

Although we performed a preliminary analysis of word histograms and
identified key differences between ZD and HF AI documentation in
specific versus abstract descriptions of key AI model attributes, we
need human-in-the-loop feedback to update the ZD templates. In future
work, we plan to investigate patterns in word histograms and
co-occurrence histograms during iterative updates of community-driven
templates for AI documentation with best practices. Furthermore, we will explore 
automated updates of past AI documentation files to keep them up-to-date with the AI advancements and 
improve reusability of AI models for a longer period of time. 

\section*{Disclaimer}
\label{sec:disclaimer}

These opinions, recommendations, findings, and conclusions do not
necessarily reflect the views or policies of NIST or the United States
Government. The code developed for this effort can be found at
\url{https://github.com/usnistgov/AI\_eval\_cards}

\section*{Acknowledgment}
\label{sec:acknowledgment}

We would like to acknowledge discussions with Michael
Majurski, Derek Juba, Razvan Amironiesen, Jesse Dunietz, and Craig
Greenberg from NIST.

\bibliographystyle{IEEEtran}
\bibliography{ai_doc}

@techreport{gil2025ai,
  author = {Yolanda Gil and Raymond Perrault},
  title = {Artificial Intelligence Index Report 2025},
  year = {2025},
  url = {https://hai.stanford.edu/assets/files/hai_ai_index_report_2025.pdf}
}

@article{brodheim2025shaping,
  author = {Ted Brodheim},
  title = {Shaping the Future of Learning: AI in Higher Education},
  year = {2025},
  month = {May},
  url = {https://er.educause.edu/articles/sponsored/2025/5/shaping-the-future-of-learning-ai-in-higher-education}
}

@misc{fda2025ai,
  author = {{FDA}},
  title = {Artificial Intelligence-Enabled Medical Devices},
  year = {2025},
  url = {https://www.fda.gov/medical-devices/software-medical-device-samd/artificial-intelligence-enabled-medical-devices}
}

@misc{onnx2019,
  author = {{Open Neural Network Exchange}},
  title = {ONNX},
  year = {2019},
  url = {https://onnx.ai/}
}

@techreport{nist2025zero,
  author = {{NIST AI Consortium}},
  title = {Extended Outline: Proposed Zero Draft for a Standard on Documentation of AI Datasets and AI Models},
  year = {2025},
  month = {September},
  url = {https://www.nist.gov/artificial-intelligence/ai-research/nists-ai-standards-zero-drafts-pilot-project-accelerate}
}

@misc{huggingface2026repo,
  author = {{Hugging Face Team}},
  title = {AI Model Repository},
  year = {2026},
  url = {https://huggingface.co/models}
}

@article{gebru2021datasheets,
  author = {Timnit Gebru and Jamie Morgenstern and Briana Vecchione and Jennifer Wortman Vaughan and Hanna Wallach and Hal Daumé III and Kate Crawford},
  title = {Datasheets for datasets},
  journal = {arXiv preprint arXiv:1803.09010},
  year = {2021}
}

@article{holland2018dataset,
  author = {Sarah Holland and Ahmed Hosny and Sarah Newman and Joshua Joseph and Kasia Chmielinski},
  title = {The dataset nutrition label: A framework to drive higher data quality standards},
  journal = {arXiv preprint arXiv:1805.03677},
  year = {2018}
}

@inproceedings{mitchell2019model,
  author = {Margaret Mitchell and others},
  title = {Model cards for model reporting},
  booktitle = {Proceedings of the Conference on Fairness, Accountability, and Transparency},
  year = {2019},
  month = {January}
}

@misc{huggingface2026cards,
  author = {{Hugging Face Team}},
  title = {Model Cards},
  year = {2026},
  url = {https://huggingface.co/docs/hub/en/model-cards}
}

@misc{google2026model,
  author = {{Google Team}},
  title = {Model Garden on Vertex AI: curated set of 200+ available models},
  year = {2026},
  url = {https://cloud.google.com/model-garden}
}

@misc{kaggle2026models,
  author = {{Kaggle Team}},
  title = {Kaggle Models},
  year = {2026},
  url = {https://www.kaggle.com/models}
}

@misc{facebook2026llama,
  author = {{Facebook Team}},
  title = {Llama models},
  year = {2026},
  url = {https://ai.meta.com/resources/models-and-libraries/}
}

@inproceedings{bajcsy2022characterization,
  author = {Peter Bajcsy and Michael Majurski and T. E. Cleveland IV and M. Carrasco and Walid Keyrouz},
  title = {Characterization of AI Model Configurations for Model Reuse},
  booktitle = {Computer Vision -- ECCV 2022 Workshops},
  year = {2022},
  pages = {454--469},
  publisher = {Springer-Verlag}
}

@misc{nltk2026,
  author = {{NLTK Team}},
  title = {Natural Language Toolkit (NLTK)},
  year = {2026},
  url = {https://www.nltk.org/}
}

@incollection{yao2023quality,
  author = {Y. Yao and H. Zhang},
  title = {Chapter Two - Quality assessment for big mobility data},
  booktitle = {Handbook of Mobility Data Mining},
  publisher = {Elsevier},
  year = {2023},
  pages = {15--34}
}

@article{behara2020novel,
  author = {K. N. Behara and A. Bhaskar and E. Chung},
  title = {A novel approach for the structural comparison of origin-destination matrices: Levenshtein distance},
  journal = {Transportation Research Part C: Emerging Technologies},
  volume = {111},
  pages = {513--530},
  year = {2020}
}

@misc{OED2026,
  author = {{Word Lucky}},
  title = {200 Words That Containing Most Letter X},
  year = {2026},
  url = {https://www.wordlucky.com/words-with-most/x?}
}


\appendix

\section{Datasets}
\label{appendix:datasets}

\begin{figure}[t]
\centerline{\includegraphics[width=37pc]{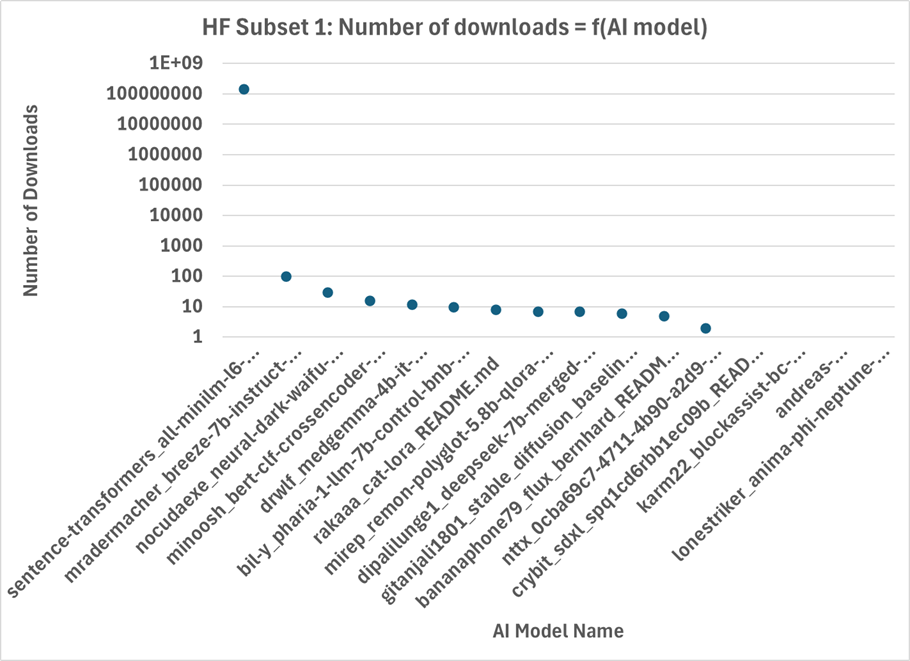}}
\caption{Number of downloads for the AI models in HF Subset 1.}
\label{fig:downloadsHFSet1}

\centerline{\rule{0.6667\linewidth}{.2pt}}
\end{figure}

Figure~\ref{fig:downloadsHFSet1} shows the number of downloads in HF Subset 1. This subset is created by sorting all AI models by the number of downloads and uniformly sampling them to represent diverse AI model reuse.

\begin{figure}[t]
\centerline{\includegraphics[width=37pc]{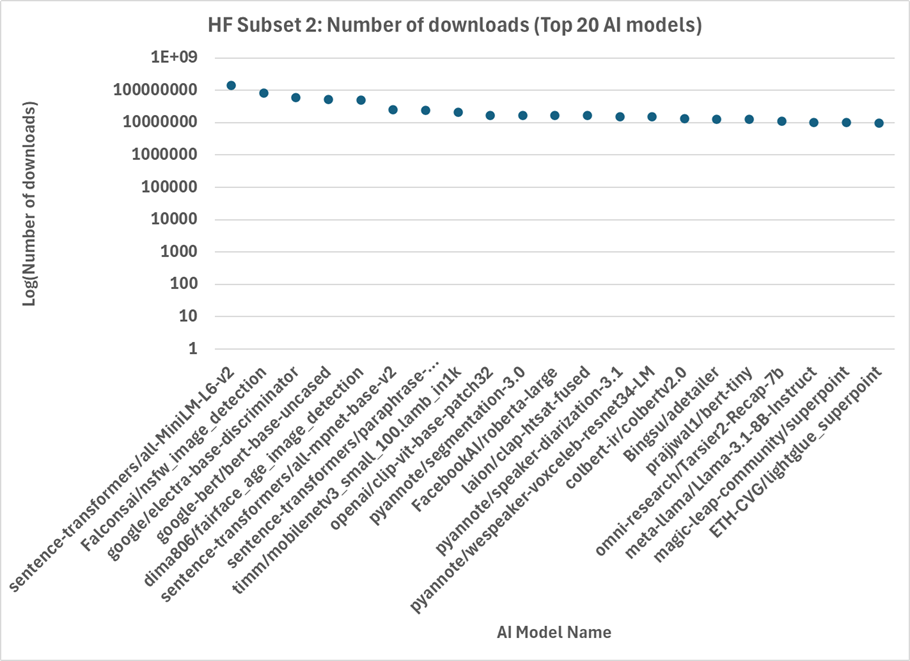}}
\caption{Number of downloads for the AI models in HF Subset 2.}
\label{fig:downloadsHFSet2}

\centerline{\rule{0.6667\linewidth}{.2pt}}
\end{figure}

Figure~\ref{fig:downloadsHFSet2} shows the number of downloads in HF Subset 2. This subset is for analyzing the most reused AI models in HF repository.

\begin{figure}[t]
\centerline{\includegraphics[width=37pc]{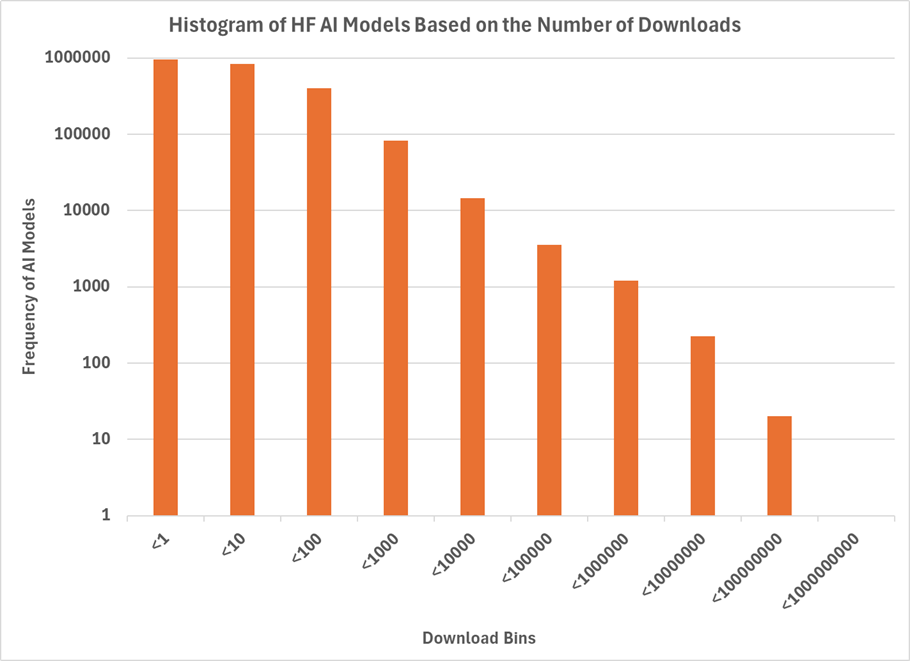}}
\caption{Visualizations of a download-based histogram of all HF AI models with increasing bin width by a factor of 10.}
\label{fig:histdownloads}

\centerline{\rule{0.6667\linewidth}{.2pt}}
\end{figure}

\begin{figure}[t]
\centerline{\includegraphics[width=37pc]{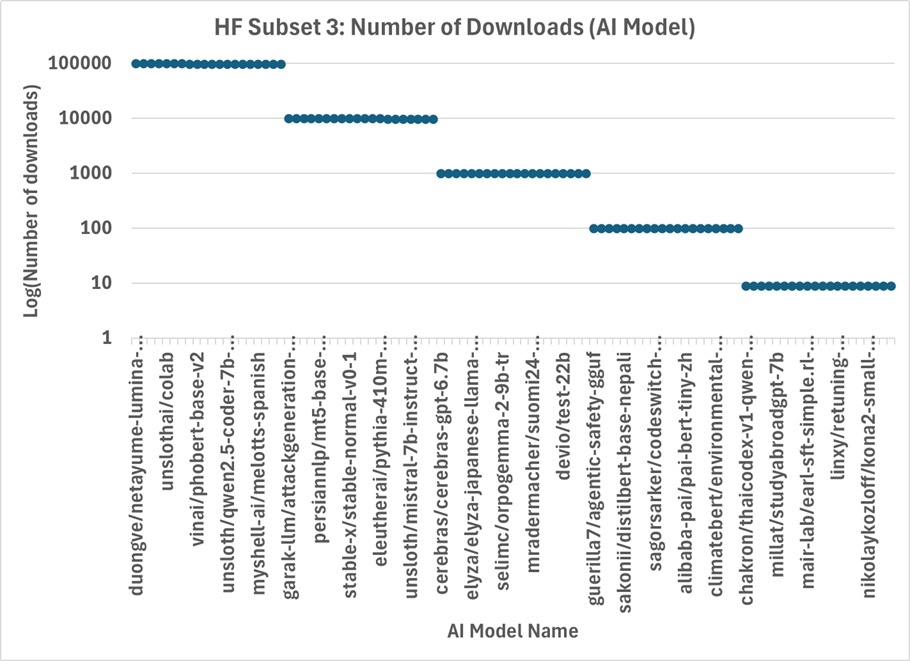}}
\caption{Number of downloads for the AI models in HF Subset 3.}
\label{fig:downloadsHFSet3}

\centerline{\rule{0.6667\linewidth}{.2pt}}
\end{figure}

Figure~\ref{fig:histdownloads} is the basis for cluster-based sampling of HF AI models and creating the HF Subset 3 of README.md files. 
This subset is created by binning all AI models based on the number of downloads, excluding bins that contain AI models with zero downloads and more than $10^5$ downloads, and then sampling the bins with 20 examples, as shown in Figure~\ref{fig:downloadsHFSet3}.

\section{Tree Visualization}
\label{appendix:toc}

\begin{figure}[t]
\centerline{\includegraphics[width=37pc]{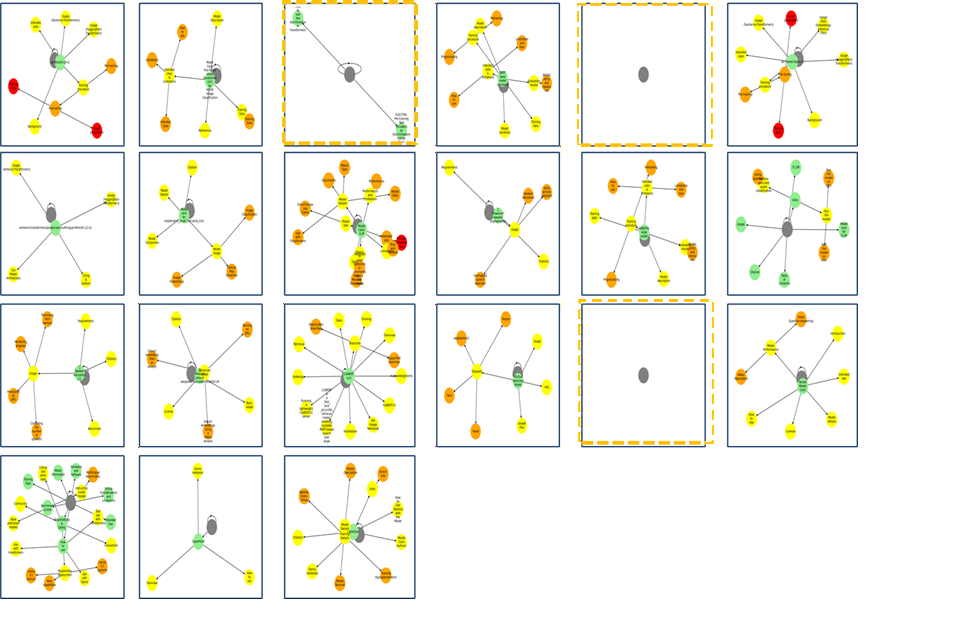}}
\caption{Tree visualization of TOCs extracted from the 21 most downloaded HF AI model cards.}
\label{fig:treesubset2}

\centerline{\rule{0.6667\linewidth}{.2pt}}
\end{figure}

Figure~\ref{fig:treesubset2} shows visualizations of 21 TOCs (table of contents) in HF Subset 2. 
These TOCs correspond to the most downloaded AI models. We highlighted three TOC visualizations with a dashed line because the TOC is minimal. 
We hypothesize that these three AI models ($google-electra-base-discriminator, dima806-fairface-age-image_detection, prajjwal1-bert-tiny$) became popular via word of mouth.

\begin{figure}[t]
\centerline{\includegraphics[width=37pc]{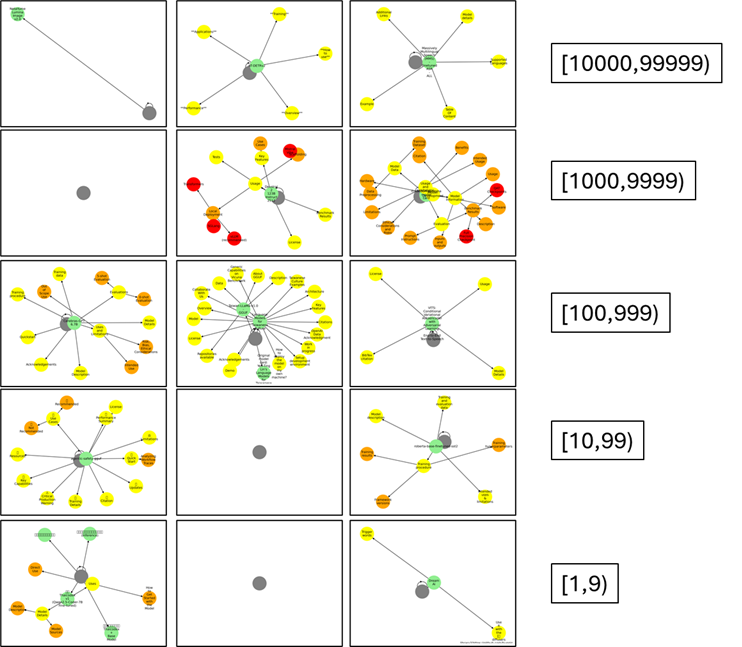}}
\caption{Three examples of TOC tree visualization from each cluster formed by binning AI models based on the number of downloads.}
\label{fig:treesubset3}
\centerline{\rule{0.6667\linewidth}{.2pt}}
\end{figure}

Figure~\ref{fig:treesubset3} shows three visualization examples of the table of contents (TOC) per cluster-based sampling used for creating HF Subset 3 (see also Figure~\ref{fig:downloadsHFSet3}). The cluster bins are shown on the right side of the figure. 
The color mapping shows green, yellow, orange, and red as heading levels 1, 2, 3, and 4 in this order. These figures illustrate the varying complexity of AI model card TOCs across AI model clusters. It is assumed that, while the HF repository is for AI model sharing, some researchers may use it as a version control system, thereby diminishing the value of the monitored download count per AI model.

\section{Examples of Similarity Scores for Matching TOCs}
\label{appendix:similarity}

Tables~\ref{table:nlss_similarity} and \ref{table:nld_similarity} show an example of TOC heading matches for the HF AI
model card named ``bil-y\_pharia-1-llm-7b-control-bnb-4bit'' being
compared to the ZD train data template. The HF AI model has been
downloaded 10 times, and its TOC tree is shown in Figure~\ref{fig:tochfsubset1} (bottom row,
2\textsuperscript{nd} from left). The NLSS-based matching rewards the
use of exactly matching informative words according to the ZD template,
with five headings matched (see Table~\ref{table:nlss_similarity}). The NLD-based matching finds
only two heading matches (see Table~\ref{table:nld_similarity}), since the normalization in NLD
gives more weight to the verbosity of the headings than in the NLSS
score, which is normalized by the length of s(HF). In this case, NLSS
and NLD similarity metrics are useful for our understanding of the
correlation between the verbosity of headings and the relevant
information as defined in the ZD templates.

\begin{table}[t]
\centering
\caption{Examples of similarity scores for \textbf{NLSS-based} matching of ZD data template and HF AI model card for the model named ''bil-y\_pharia-1-llm-7b-control-bnb-4bit''. The match is valid for NLSS$\geq 25$.}
\label{table:nlss_similarity}
\begin{tabularx}{\columnwidth}{|>{\hsize=0.5\hsize\centering\arraybackslash}X|>{\hsize=1.5\hsize\raggedright\arraybackslash}X|>{\hsize=1.5\hsize\raggedright\arraybackslash}X|>{\hsize=1.0\hsize\raggedright\arraybackslash}X|>{\hsize=0.5\hsize\centering\arraybackslash}X|}
\hline
\textbf{Match Idx} & \textbf{Heading string from ZD data template} & \textbf{Heading string from HF AI model card} & \textbf{Common Words} & \textbf{NLSS Score} \\ \hline
1 & Appendix 1: Expanded Dataset Documentation Template Dataset Identifying Descriptors Contact \emph{\textbf{Details}} Source URL & Model Card for Model ID Model \emph{\textbf{Details}} & \{`Details'\} & 29 \\ \hline
2 & Appendix 1: Expanded Dataset Documentation Template Intended Use \emph{\textbf{Out-of-Scope Use}} & Model Card for Model ID Uses \emph{\textbf{Out-of-Scope Use}} & \{`Out-of-Scope', `Use'\} & 43 \\ \hline
3 & Appendix 1: Expanded Dataset Documentation Template Dataset \emph{\textbf{Evaluation}} Data Visualization & Model Card for Model ID \emph{\textbf{Evaluation}} & \{`Evaluation'\} & 37 \\ \hline
4 & Appendix 1: Expanded Dataset Documentation Template Dataset \emph{\textbf{Evaluation}} Data Visualization & Model Card for Model ID \emph{\textbf{Evaluation}} Results & \{`Evaluation'\} & 28 \\ \hline
5 & Appendix 1: Expanded Dataset Documentation Template Dataset Identifying Descriptors \emph{\textbf{Contact}} Details Source URL & Model Card for Model ID Model Card \emph{\textbf{Contact}} & \{`Contact'\} & 29 \\
\hline
\end{tabularx}

\vspace{2ex}
\centerline{\rule{0.6667\linewidth}{.2pt}}
\end{table}

\begin{table}[htbp]
\centering
\caption{Examples of similarity scores for \textbf{NLD-based} matching of ZD data template and HF AI model card for the model named ''bil-y\_pharia-1-llm-7b-control-bnb-4bit''. The match is valid for $NLD_{sorted} \geq 50$.}
\label{table:nld_similarity}
\begin{tabularx}{\columnwidth}{|>{\hsize=0.3\hsize\centering\arraybackslash}X|>{\hsize=1.35\hsize\raggedright\arraybackslash}X|>{\hsize=1.35\hsize\raggedright\arraybackslash}X|>{\hsize=1.0\hsize\raggedright\arraybackslash}X|>{\hsize=1.0\hsize\centering\arraybackslash}X|}
\hline
\textbf{Match Idx} & \textbf{Heading string from ZD data template} & \textbf{Heading string from HF AI model card} & \textbf{$NLD_{ratio}$ score} & \textbf{$NLD_{sorted}$ score} \\ \hline
1 & Appendix 1: Expanded Dataset Documentation Template Dataset Identifying Descriptors & Model Card for Model ID Evaluation Testing Data, Factors \& Metrics Testing Data & 44 & 51 \\ \hline
2 & Appendix 1: Expanded Dataset Documentation Template Intended Use Out-of-Scope Use & Model Card for Model ID Uses Out-of-Scope Use & 51 & 53 \\
\hline
\end{tabularx}

\vspace{2ex}
\centerline{\rule{0.6667\linewidth}{.2pt}}
\end{table}

\section{Word Histograms}
\label{appendix:hist}

\begin{figure}[t]
\centerline{\includegraphics[width=37pc]{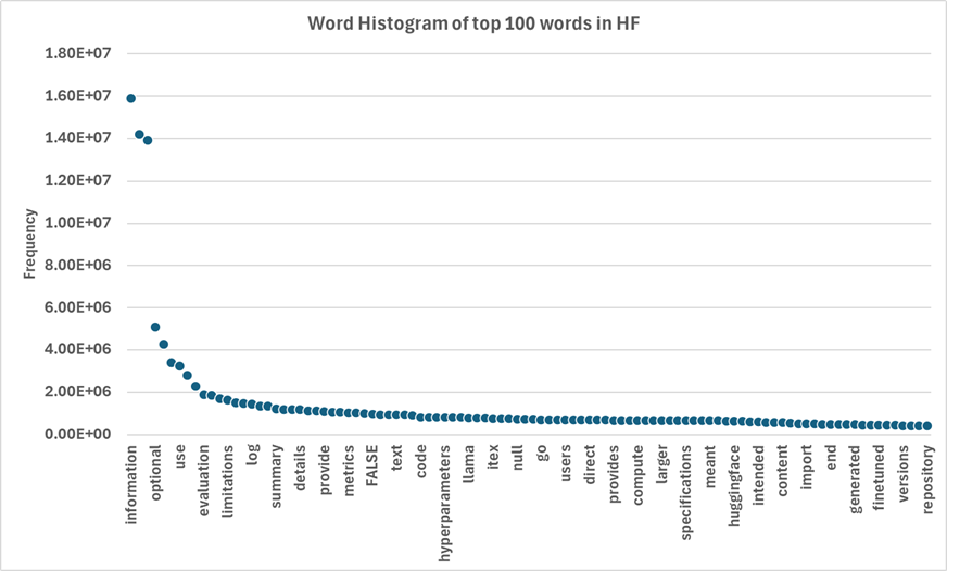}}
\caption{Histogram values of top 100 words occurring in HF AI model documentation.}
\label{fig:hfdownloads}

\centerline{\rule{0.6667\linewidth}{.2pt}}
\end{figure}

\begin{figure}[t]
\centerline{\includegraphics[width=37pc]{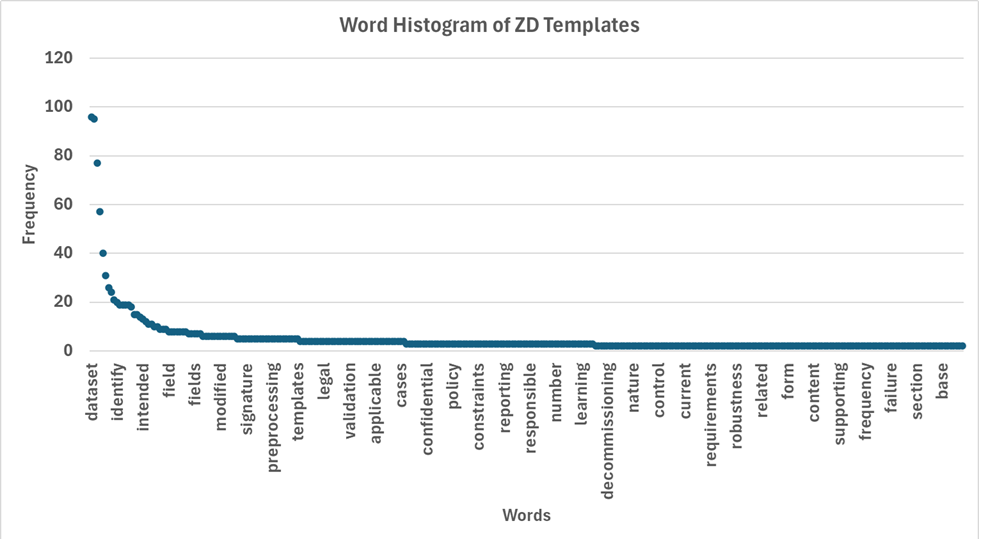}}
\caption{Histogram values of the top 100 words occurring in ZD AI templates.}
\label{fig:zddownloads}

\centerline{\rule{0.6667\linewidth}{.2pt}}
\end{figure}

We illustrate the top 100 most frequently used words in HF AI model cards in Figure~\ref{fig:hfdownloads} 
and the top 100 most frequently used words in ZD templates in Figure~\ref{fig:zddownloads}.

\section{Correlations}
\label{appendix:correlation}

\begin{figure}[t]
\centerline{\includegraphics[width=37pc]{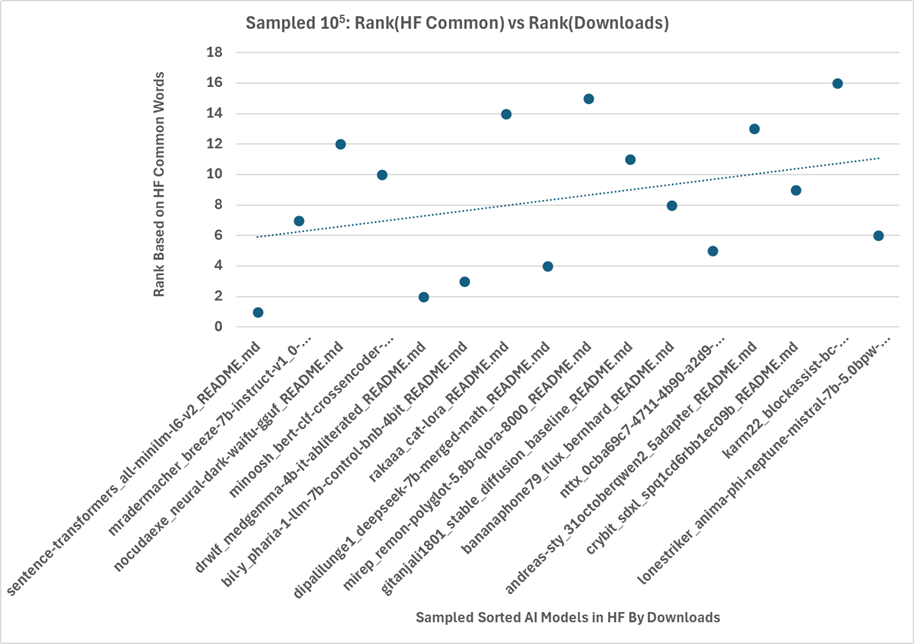}}
\caption{A scatter plot of ``Rank vs Rank'' for HF Subset 1. The ranks of README.md files in HF Subset 1 (sorted and sampled AI models according to downloads) are computed based on the common words with the HF word histogram and the number of downloads. The horizontal axis displays AI model names, ranked from the largest to the smallest number of downloads.}
\label{fig:rank2}

\centerline{\rule{0.6667\linewidth}{.2pt}}
\end{figure}

We used the common words between ZD templates and all HF AI model cards as the reference for ranking. In this appendix, we use the histogram of all HF AI model cards as the reference for ranking. 
When using the HF histogram as a reference, we observe a higher correlation for HF Subset 1 in Figure~\ref{fig:rank2}, since the reference template follows best HF practices. 
The correlation ``Rank vs Rank'' is 0.37, while the correlation ``Freq vs Freq'' is 0.44. 
The deltas in correlation values for ZD or HF histograms as references are $0.37 - 0.31 = 0.06$ for ranks and $0.44 - 0.23 = 0.21$ for frequencies, indicating some room for ZD word improvement if it becomes the standard template. 


\end{document}